\documentclass[10pt,twocolumn,letterpaper]{article}

\usepackage[pagenumbers]{cvpr} 

\usepackage{times}
\usepackage{epsfig}
\usepackage{graphicx}
\usepackage{amsmath}
\usepackage{amssymb}
\usepackage{booktabs}
\usepackage{color}
\usepackage{kotex}

\usepackage{adjustbox}
\usepackage{algorithm}
\usepackage{algorithmic}

\usepackage{multirow}

\usepackage[pagebackref,breaklinks,colorlinks]{hyperref}

\usepackage{enumitem}

\usepackage[capitalize]{cleveref}
\crefname{section}{Sec.}{Secs.}
\Crefname{section}{Section}{Sections}
\Crefname{table}{Table}{Tables}
\crefname{table}{Tab.}{Tabs.}

\def\cvprPaperID{4603} 
\def\confName{CVPR}
\def\confYear{2022}

\begin{document}

\title{MUM : Mix Image Tiles and UnMix Feature Tiles \\ for Semi-Supervised Object Detection}

\author{
JongMok Kim$^{1,2}$~~~
JooYoung Jang$^{1,2}$~~~
Seunghyeon Seo$^{2}$~~~
Jisoo Jeong$^{2}$~~~
Jongkeun Na$^{1}$~~~
Nojun Kwak$^{2}$~~~
\smallskip
\\
$^1$SNUAILAB~~~
$^2$Seoul National University
}

\maketitle

\begin{abstract}
Many recent semi-supervised learning (SSL) studies build teacher-student architecture and train the student network by the generated supervisory signal from the teacher.
Data augmentation strategy plays a significant role in the SSL framework since it is hard to create a weak-strong augmented input pair without losing label information.
Especially when extending SSL to semi-supervised object detection (SSOD), many strong augmentation methodologies related to image geometry and interpolation-regularization are hard to utilize since they possibly hurt the location information of the bounding box in the object detection task. 
To address this, we introduce a simple yet effective data augmentation method, Mix/UnMix (MUM), 
which unmixes feature tiles for the mixed image tiles for the SSOD framework.
Our proposed method makes mixed input image tiles and reconstructs them in the feature space.
Thus, MUM can enjoy the interpolation-regularization effect from non-interpolated pseudo-labels and successfully generate a meaningful weak-strong pair.
Furthermore, MUM can be easily equipped on top of various SSOD methods.
Extensive experiments on MS-COCO and PASCAL VOC datasets demonstrate the superiority of MUM by consistently improving the mAP performance over the baseline in all the tested SSOD benchmark protocols. 
The code is released at \href{https://github.com/JongMokKim/mix-unmix}{https://github.com/JongMokKim/mix-unmix.}

\end{abstract}

\section{Introduction}
\label{sec:intro}

Deep neural networks have made a lot of progress on diverse computer vision tasks thanks to the availability of large-scale datasets.
To achieve better and generalizable performance, a large amount of labeled data is indispensable, which however requires a vast amount of workforce and time for annotation\cite{russakovsky2015best, bearman2016s, dollar2012pedestrian}.   
Unlike image classification, which needs only a class label per image, object detection requires a pair of a class label and location information for multiple objects per single image. 
Therefore, it is more challenging to acquire enough amount of labeled data in object detection. 
To address the above problem, many recent works have focused on leveraging abundant unlabeled data when training the network with a small amount of labeled data, called \textit{semi-supervised learning} (SSL) and \textit{semi-supervised object detection} (SSOD).

\begin{figure} 
\includegraphics[width=1.0\linewidth]{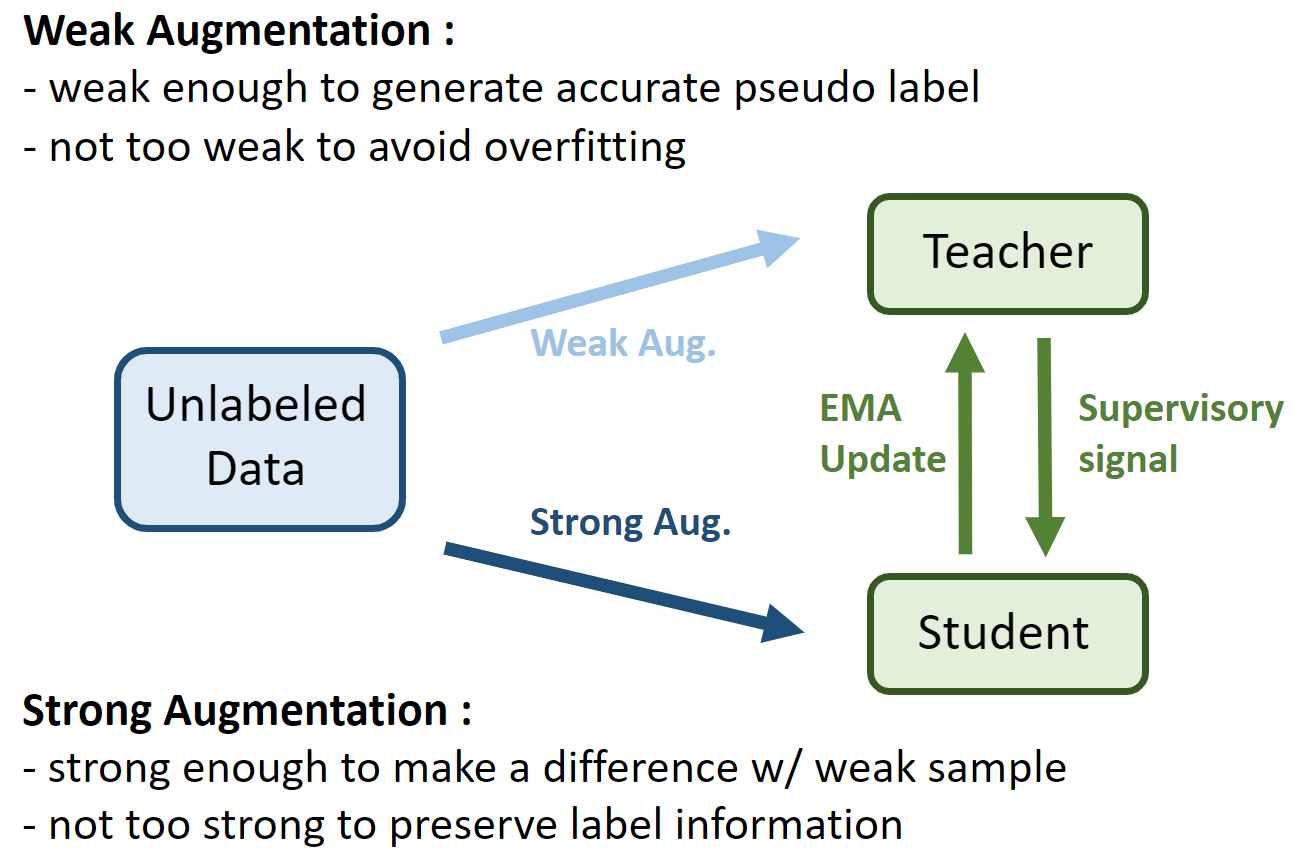} 
\vspace{-5mm}
\caption{\textbf{Typical teacher-student (pseudo-labeling) framework for SSL.} 
To fully exploit the unlabeled data, building an intelligent teacher and employing an adequate data augmentation strategy for weak-strong pairs are very important in this framework.
}

\label{fig:ssl_system}
\vspace{-5mm}
\end{figure} 

\begin{figure*}[t] 
\begin{center}
\includegraphics[width=1.0\linewidth]{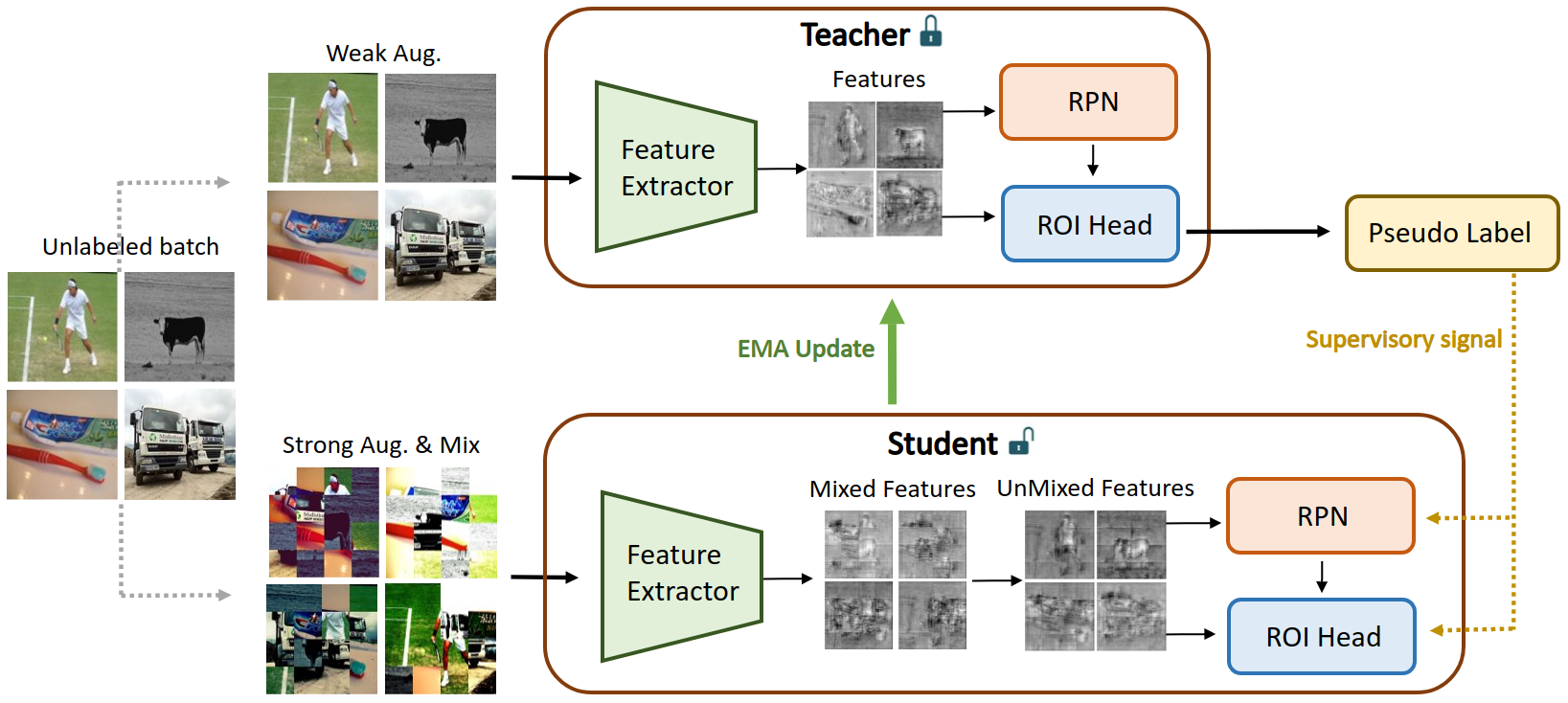}
\end{center}
\caption{\textbf{Overview of Mix/UnMix (MUM) training system.} The teacher network generates a pseudo label to give a supervisory signal to the student, while weakly and strong \& mixed augmented inputs are injected to the teacher and the student, respectively. In order to utilize the supervisory signal from the original shaped image, we unmix the mixed feature tiles and feed the unmixed features to the detection head in the student network. In each training step, the teacher network is slowly updated via EMA of the student's weights. For visual simplicity, we assume the batch size, $N_T$, and $N_G$ are all identical to 4. For more details about the hyperparameters, $N_T$ and $N_G$, see Sec.\ref{sec:method}.}
\label{fig:tut_overview}
\end{figure*} 

In recent days, many SSL works rely on the teacher-student framework where a teacher network, typically a temporal ensemble model of the student, generates supervisory signals and trains the student network with them as shown in Fig. \ref{fig:ssl_system} \cite{laine2016temporal, tarvainen2017mean}.
Data augmentation plays a significant role in this framework and most of the recent works apply strongly augmented inputs for the student model while weak augmentations are given to the teacher \cite{verma2019interpolation, sohn2020fixmatch}. 
Interpolation-regularization (IR), whose core idea is that the output of the interpolated input should be similar to the interpolated output of the original inputs, was originally developed as a data augmentation technique for supervised learning \cite{zhang2017mixup} and has been successfully applied to teacher-student framework for SSL \cite{verma2019interpolation, berthelot2019mixmatch}. It is a clever way to generate augmented input-output pairs without losing much contextual information and has also been extended to semantic segmentation by generating interpolated labels in a pixel-wise manner \cite{french2019semi,kim2020structured}.

However, it is challenging to make an interpolated label in the object detection task because it involves in multi-task learning, which consists of localization and classification.   
To tackle this problem, in this paper, we propose Mix/UnMix (MUM) method, which exploits IR in a much more efficient and more straightforward way for object detection (Fig. \ref{fig:tut_overview}).
MUM generates mixed images\footnote{Mixed images can be considered as a type of interpolated images since they can be generated by patchwise interpolation with binary interpolation coefficients.} by mixing image tiles in a batch and uses them as inputs to the student network.
Then the feature maps extracted from the backbone are unmixed back to their original image geometry.
The tiles maintain their positions in the original images through the mixing process so that it is possible for the feature maps to get back to their initial position through an unmixing phase.
Therefore, the student network can learn from the mixed image without the interpolated (mixed) label.
For the teacher network, input images are weakly augmented to generate highly confident pseudo labels as in other existing methods.
As a result, the student can learn the robust features from the mixed and naturally occluded input image with the guide of a confident pseudo-label from the teacher network. 

We benchmark Unbiased-Teacher\cite{liu2021unbiased} as a reliable baseline, which proposed a pseudo-labeling method for SSOD.
Following the standard experimental setting of recent SSOD research, we adopted Faster-RCNN\cite{ren2015faster} as a default architecture.
To verify the superiority of our algorithm, we test MUM on PASCAL VOC\cite{everingham2010pascal} and MS-COCO\cite{lin2014microsoft} dataset following the experimental protocols used in \cite{liu2021unbiased}.
MUM achieves performance improvement against the baseline in every experimental protocol and could obtain the state-of-the-art performance in the SSOD benchmark experiment.
Furthermore, thanks to the simplicity of MUM, the increase in the computational cost and complexity is negligible in the train phase, 
and it can be readily plugged in other SSOD frameworks as a data augmentation method.
We also explore the versatility of MUM over different architectures through additional experiments with the Swin Transformer backbone.
In addition, we tested performance of MUM for the supervised ImageNet classification task \cite{deng2009imagenet}.
Our main contributions can be summarized as follows:

\begin{itemize}[leftmargin=*]
\vspace{-2mm}
\item We show the problem in applying the IR method to the pseudo-label-based semi-supervised object detection and propose a novel and simple data augmentation method, MUM, which benefits from IR.
\vspace{-2mm}
\item We experimentally prove our proposed method's superiority over a reliable baseline method through experiments and could obtain state-of-the-art performance on MS-COCO and PASCAL VOC dataset. 
Furthermore, we demonstrate the generalizability of our proposed method by still getting improved performance on a different backbone, Swin Transformer.
\vspace{-2mm}
\item Through thorough analysis of the feature maps, class activation maps and experimental results, we show the proposed MUM's compatibility with the SSOD problem.
\end{itemize}

\section{Related Work}
\label{sec:related}

\subsection{Semi-Supervised Learning}
Since semi-supervised learning tackles practical problems regarding the cost of labeling and raw data acquisition, considerable progress has been made on improving the performance using only a few labeled data in combination with plenty of unlabeled data. 
Most SSL methods can be classified into two categories according how to generate supervisory signals from unlabeled raw data: consistency-based method~\cite{berthelot2019mixmatch, berthelot2019remixmatch, laine2016temporal, tarvainen2017mean, miyato2018virtual, xie2019unsupervised, verma2019interpolation} which induces consistent predictions for the same but differently augmented images and pseudo-labeling method\cite{lee2013pseudo, bachman2014learning, iscen2019label, arazo2020pseudo, xie2020self, sohn2020fixmatch} which trains a student network with the highly confident label from a teacher network.
As shown in Fig.\ref{fig:ssl_system}, to generate meaningful supervisory signals in the pseudo-labeling approach, it is necessary to equip with both a good teacher which can make better predictions than a student and an effective data augmentation method for generating data with different levels of difficulty under the same label. 
The most common and efficient method of building the teacher network is \textit{Exponential Moving Average} (EMA)\cite{tarvainen2017mean} which updates the teacher with a temporal ensemble of the student network.
Regarding data augmentation, UDA~\cite{xie2019unsupervised}, ReMixMatch~\cite{berthelot2019remixmatch} and FixMatch~\cite{sohn2020fixmatch} apply RandAugment~\cite{cubuk2019randaugment}, CTAugment~\cite{berthelot2019remixmatch} and Cutout~\cite{devries2017cutout} as strong augmentations to generate data more difficult to learn than those from weak augmentations to make more meaningful supervisory signals.
Interpolated-regularization is one of the efficient data augmentation methods in SSL and will be discussed further in Sec.\ref{sec:related_ir}.

\subsection{Semi-Supervised Object Detection}
SSOD has gained significant attention for reducing burdensome cost of labeling in object detection task \cite{wang2021data, yang2021interactive, zhou2021instant, sohn2020simple, Jeong_2021_CVPR, jeong2019consistency, liu2021unbiased}. 
CSD\cite{jeong2019consistency} applied the consistency-regularization method, one of the mainstreams in SSL, into the object detection task. STAC\cite{sohn2020simple} proposed the simple framework that trains a student network with pseudo-labels generated by a fixed teacher using unlabeled data.
However, the fixed teacher network trained with only labeled data is insufficient to generate enough reliable pseudo-labels.

A line of recent works improves the teacher network and its pseudo-label by multi-phase training\cite{wang2021data} or updates the teacher online by EMA\cite{liu2021unbiased, yang2021interactive, zhou2021instant}, similar to MeanTeacher\cite{tarvainen2017mean}.
It leads to a reciprocal structure so that a teacher network generates supervisory signals helpful for improving the performance of a student network, and the teacher can also be strengthened by EMA update.
Unbiased-Teacher\cite{liu2021unbiased} is composed of a simple SSOD framework that is robust to error propagation, using existing techniques such as EMA and Focal Loss\cite{lin2017focal}. 
It also made use of both strong and weak data augmentation, similar to FixMatch\cite{sohn2020fixmatch}.

In contrast to SSL for classification tasks, the data augmentation methods in SSOD require the geometry of each augmented image to be identical for utilizing localization information from the teacher network’s output as a supervisory signal.
To overcome this constraint, we propose MUM that can mutate image geometry diversely and reduce the error propagation drastically.

\subsection{Interpolation-based Regularization}
\label{sec:related_ir}
IR is a method that derives high performance of a deep learning network by preprocessing inputs without noise injection and has been actively studied until recently \cite{ge2021yolox, verma2018manifold, zhang2017mixup, tokozume2017betweenclass, cutmix, devries2017cutout,acai, beckham2019adversarial}.
It generates new training samples by interpolating the original ones based on the inductive bias; the linear combination of two original samples’ outputs should be similar to the output of the interpolated sample.
Mixup\cite{zhang2017mixup}, CutMix\cite{cutmix}, Mosaic\cite{ge2021yolox}, and Cutout\cite{devries2017cutout} are methods to synthesize and generate training samples and Manifold Mixup\cite{verma2018manifold} deals with hidden representations in the feature level rather than with original images. 
Such methods can be regarded as strong data augmentation, and there have been several attempts to utilize them in SSL and SSOD. 

ICT\cite{verma2019interpolation} trains a network by consistency loss between the interpolated prediction of two unlabeled samples and the prediction of an interpolated sample.
MixMatch\cite{berthelot2019mixmatch} and ReMixMatch\cite{berthelot2019remixmatch} generate a guessed label from multi-view of a single unlabeled image, and then train it via Mixup\cite{zhang2017mixup} with labeled training sample.
In addition, \cite{french2019semi, kim2020structured} extends SSL to the semantic segmentation by generating mixed images via CutMix\cite{cutmix} and training with the same mechanism as ICT\cite{verma2019interpolation}.

Unbiased-Teacher\cite{liu2021unbiased} also used Cutout\cite{devries2017cutout} as a strong augmentation.
However, Cutout results in information loss to the inputs because it drops pixel values of the random box-shape area in an image.
Although ISD\cite{Jeong_2021_CVPR} applied IR adequately into the SSOD framework, it can be categorized more as a consistency-based method.
Instant-Teaching\cite{zhou2021instant} applied Mixup\cite{zhang2017mixup} and Mosaic\cite{ge2021yolox} directly into pseudo-label-based SSOD framework, but the problem of class ambiguity of mixture between backgrounds and objects remains unsolved as mentioned in ISD\cite{Jeong_2021_CVPR}.
To summarize, while Cutout\cite{devries2017cutout} has a weak regularization effect,
Mixup\cite{zhang2017mixup} has the class ambiguity issue in the interpolated label generation process.
Motivated by these limitations, we propose MUM not only to avoid the problem caused by interpolated labels but also to still enjoy the IR effect. 

\begin{figure*}[ht] 
\includegraphics[width=1.\linewidth]{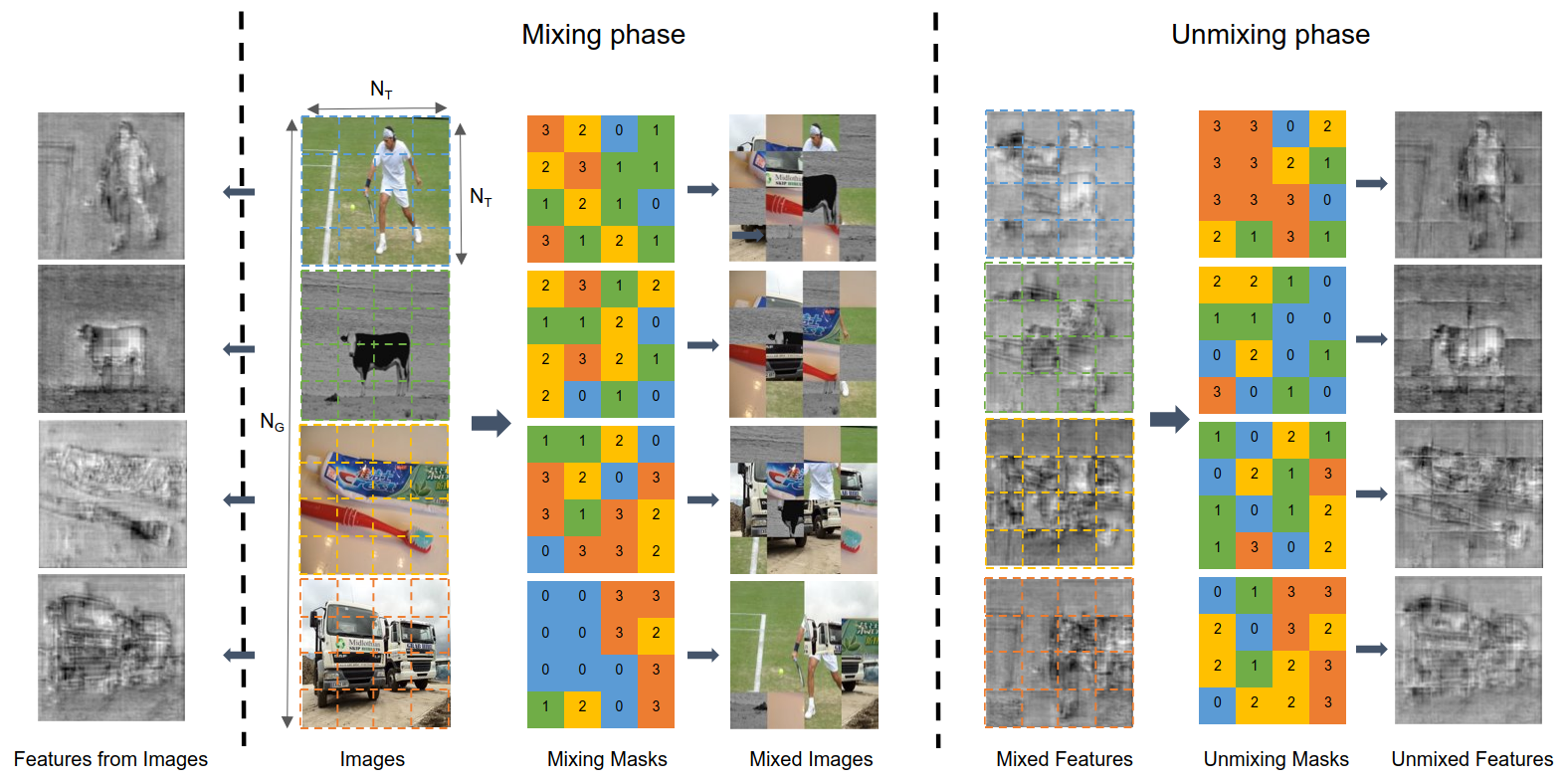}
\caption{We provide the detailed operation of MUM with enlarged figures of images and features in Fig.\ref{fig:tut_overview}.
With an assumption of $N_G = N_T = 4$, 4 images form a group, and each image is split into 4$\times$4 tiles.
Next, each input tile is mapped to the mixed image in the corresponding position of each mixing mask.
Similar to the mixing phase, unmixed features are generated from mixed features with unmixing masks.
Note that we generate mixing masks stochastically in each training step and unmixing masks are made from the mixing masks.
Additionally, we provide the features from the original images to compare with unmixed features.
}
\label{fig:tile_untile_feat}
\end{figure*} 

\section{Method}
\label{sec:method}

\subsection{Preliminary}
\label{subsec:problem}
\noindent \textbf{Problem definition.} We deal with the semi-supervised object detection task, where a set of labeled data $\mathbf{D_s}=\{(x^s_i,y^s_i)\}^{N_s}_{i=1}$ and unlabeled data $\mathbf{D_u}=\{x^u_j\}^{N_u}_{j=1}$ is given for training. 
Here, $x, y, N_s, N_u$ denote an image, the corresponding label, the number of labeled and unlabeled samples, respectively.

\noindent \textbf{Baseline.} Unbiased-teacher\cite{liu2021unbiased}  is a well-designed architecture that employs the existing but competitive techniques like the Focal loss and EMA update method.
They build a stable SSOD system with an unbiased teacher and its confident pseudo labels. 
To keep the above benefits, we choose it as our baseline.
Following the baseline, we first build the teacher network via EMA:
\begin{equation}\label{eq1}
\theta^{t+1} = \theta^t \cdot \delta + \theta \cdot (1-\delta),
\end{equation}
where $\theta^t, \theta$ and $\delta$ denote the weights of the teacher at step $t$, the weights of the student, and EMA decay rate, respectively. 
Since the model performance is sensitive to the decay rate $\delta$, it is essential to set the proper value to make the teacher better than the student.
We will further discuss the effect of the decay rate $\delta$ on the system performance in Sec.\ref{sec:discussions}.

Next, we train the student network with the pseudo labels generated by the teacher network.
The total training loss, $\mathcal{L}$, consisting of the supervised loss, $\mathcal{L}_s$, and the unsupervised loss, $\mathcal{L}_u$, can be described as follows:
\begin{equation}\label{eq2}
\begin{split}
\mathcal{L}_s &= \sum_i{\mathcal{L}_{cls}(x^s_i,y^s_i)+\mathcal{L}_{reg}(x^s_i,y^s_i)} ,\\
\mathcal{L}_u &= \sum_i{\mathcal{L}_{cls}(x^u_i,\hat{y}^u_i)+\mathcal{L}_{reg}(x^u_i,\hat{y}^u_i)}, \\
\mathcal{L} &= \mathcal{L}_s + \lambda_u \cdot \mathcal{L}_u,
\end{split}
\end{equation}
where $\mathcal{L}_{cls}, \mathcal{L}_{reg}, \hat{y}^u$ and $\lambda_u$ denote the loss for classification, the loss for bounding box regression, a pseudo label for an unlabeled image given by the teacher, and the balancing weight for the unsupervised loss, respectively. 

\begin{algorithm}[t]
\caption{Training procedure of the proposed MUM }
\label{algorithm:tut}
\textbf{Require}: $(\mathcal{X}^s,\mathcal{Y}^s), \mathcal{X}^u$: pair of images and its labels, and unlabeled images
\\\textbf{Require}: $h(\cdot), \lambda_u$: loss function and balancing weight
\\\textbf{Require}: $f_{b,t}(\cdot)$, $f_{d,t}(\cdot)$: teacher object detection model (backbone and detector network) 
\\\textbf{Require}: $f_{b,s}(\cdot)$, $f_{d,s}(\cdot)$: student object detection model (backbone and detector network)
\\\textbf{Require}: $m(\cdot)$, $u(\cdot)$: mixing and unmixing function
\\\textbf{Require}: $w(\cdot)$, $s(\cdot)$: weak and strong augmentation

\begin{algorithmic}[1]
\FOR {each $t \in [1$, max\_iterations$]$}
\STATE \textit{\textbf{Prepare Data}}
\STATE \quad $\mathcal{A} \gets w(\mathcal{X}^s)+s(\mathcal{X}^s)$, 
$\mathcal{B} \gets w(\mathcal{X}^u)$, \ $\mathcal{C} \gets s(\mathcal{X}^u)$\\
\STATE \textit{\textbf{Compute the supervised loss}}
\STATE \quad $ \mathcal{P}^{s} \gets f_{d,s}(f_{b,s}(\mathcal{A})) $
\STATE \quad $ \mathcal{L_{S}} \gets h(\mathcal{P}^{s},\mathcal{Y}^s) $
\STATE \textit{\textbf{Generate Pseudo Label}}
\STATE \quad $ \mathcal{\hat{Y}}^u \gets f_{d,t}(f_{b,t}(\mathcal{B})) $
\STATE \textit{\textbf{Mix Image Tiles \& Unmix Feature Tiles}}
\STATE \quad $fm \gets u(f_{b,s}(m(\mathcal{C})))$ 
\STATE \textit{\textbf{Compute the unsupervised loss}}
\STATE \quad  $\mathcal{P}^{u} \gets f_{d,s}(fm)$
\STATE \quad  $\mathcal{L_{U}} \gets h(\mathcal{P}^{u},\mathcal{\hat{Y}}^u)$
\STATE \textit{\textbf{Compute the total loss}}
\STATE \quad $ \mathcal{L}_{Total} \gets \mathcal{L_{S}} + \lambda_u\cdot\mathcal{L_U} $
\STATE \textit{\textbf{Update $f_{s}(\cdot)$}} via $\mathcal{L}_{Total}$ and \textbf{$f_{t}(\cdot)$} via EMA
\ENDFOR
\end{algorithmic}
\end{algorithm}

\subsection{Mixing Image/Unmixing Feature (MUM)}
\noindent\textbf{MUM.} This section introduces the competitive data augmentation strategy, MUM (Mixing image tiles and UnMixing feature tiles), to leverage the unlabeled data effectively.
Similar to the previous IR methods such as Mixup\cite{zhang2017mixup} and CutMix\cite{cutmix}, we generate interpolated samples from each input mini-batch. 
We first split each image into $N_T \times N_T$ tiles. 
Simultaneously, we generate the same shaped $N_T \times N_T$ mask to mix each image tile and get each feature tile back to its original position.
Note that in the mixing phase all image tiles should be used once and keep their original geometric position in the image space for future reconstruction in the unmixing phase.
In order to avoid the effect of mini-batch size on mixing, we predefine the number of images to form a group to mix as $N_G$.
For example, assuming the mini-batch size of 32 and $N_G = 4$, then it forms 8 groups and the images are tiled and mixed within the corresponding group.
The detailed example of MUM operation is provided in Fig.\ref{fig:tile_untile_feat}.

Even though mixing tiles makes it hard to identify the edge or part of objects in images and feature maps, unmixing recovers the original position of features without loss of information.
Unmixed features look degraded than the features from the original image since mixing tiles incurs severe occlusion so that each piece of feature tile can only make use of its local information.
Therefore, MUM makes the student endeavor to predict like the teacher even with weak clues in features, and it is in line with the philosophy of previous studies\cite{sohn2020fixmatch, xie2019unsupervised, kim2020structured} about weak-strong data augmentations.

\noindent\textbf{Overall SSOD Framework.} Employing MUM, we design the SSOD framework as shown in Fig.\ref{fig:tut_overview}.
Similar to the baseline, we build the SSOD framework upon the pseudo-labeling method and the proposed MUM data augmentation.
A mini-batch of unlabeled images is applied to the weak and strong augmentation as inputs to the teacher and student networks. 
The methods used to generate weak and strong augmentations are identical to those for the baseline\cite{liu2021unbiased}.
Additionally for the student, we split and mix the input image tiles to generate mixed inputs and the feature maps of the mixed images are generated by the feature extractor. 
Then the mixed features are unmixed so that the original positions of all the tiles are restored.
On the other hand, the teacher generates the supervisory signal for the inputs without the mixing process.
Note that MUM can achieve the interpolation-regularization effect with a pseudo label of a single image because of the mixing-unmixing process in the student network.
Including the above unsupervised learning process, the whole training process is described in Algorithm.\ref{algorithm:tut}.


\begin{table*}[ht]
\caption{Experimental results ($AP_{50:95}$) on MS-COCO dataset with COCO-Standard and COCO-Additional protocols. 
}
\centering
\begin{adjustbox}{width=0.9\textwidth}
\begin{tabular}{|l|ccccc|c|}
\hline
\multicolumn{1}{|c|}{\multirow{2}{*}{Methods}} & \multicolumn{5}{c|}{COCO-Standard}                                                                                                                                     & \multirow{2}{*}{COCO-Additional} \\
\multicolumn{1}{|c|}{}                         & 0.5\%                                          & 1\%                         & 2\%                         & 5\%                         & 10\%                        &                                  \\ \hline
Supservised                                    & {6.83$\pm$0.15} & 9.05$\pm$0.16  & 12.70$\pm$0.15 & 18.47$\pm$0.22 & 23.86$\pm$0.81 & 37.63                            \\
CSD\cite{jeong2019consistency}                                            & 7.41$\pm$0.21                     & 10.51$\pm$0.06 & 13.93$\pm$0.12 & 18.63$\pm$0.07 & 22.46$\pm$0.08 & 38.82                            \\
STAC\cite{sohn2020simple}                                           & 9.78$\pm$0.53                     & 13.97$\pm$0.35 & 18.25$\pm$0.25 & 24.38$\pm$0.12 & 28.64$\pm$0.21 & 39.21                            \\
Instant Teaching\cite{zhou2021instant}                               & -                                              & 18.05$\pm$0.15 & 22.45$\pm$0.15 & 26.75$\pm$0.05 & 30.40$\pm$0.05 & 39.6                             \\
ISMT\cite{yang2021interactive}                                           & -                                              & 18.88$\pm$0.74 & 22.43$\pm$0.56 & 26.27$\pm$0.24 & 30.53$\pm$0.52 & 39.6                             \\
Multi Phase\cite{wang2021data}                                    & -                                              & -                           & -                           & -                           & -                           & 40.1                             \\
Unbiased Teacher\cite{liu2021unbiased}                               & 16.94$\pm$0.23                    & 20.75$\pm$0.12 & 24.30$\pm$0.07 & 28.27$\pm$0.11 & 31.50$\pm$0.10   & 41.3                             \\ \hline
MUM(Ours)                                            & \textbf{18.54$\pm$0.48}                    & \textbf{21.88$\pm$0.12} & \textbf{24.84$\pm$0.10}  & \textbf{28.52$\pm$0.09} & \textbf{31.87$\pm$0.30}  & \textbf{42.11}                            \\ \hline
\end{tabular}
\end{adjustbox}
\label{tab:coco}
\end{table*}

\begin{table}[ht]
\caption{Experimental results on PASCAL VOC dataset compared with recent state-of-the-art results. Both protocols equally use VOC07 as labeled training dataset.} 
\centering
\adjustbox{width=0.95\linewidth}{
\begin{tabular}{|l|c|cc|}
\hline
\multicolumn{1}{|c|}{Methods} & Unlabeled                                              & $AP_{50}$     & $AP_{50:95}$  \\ \hline
Supervised                    & None                                                                           & 72.63          & 42.13          \\ \hline
CSD\cite{jeong2019consistency}                           & \multirow{7}{*}{VOC12}                                                         & 74.70          & -              \\
STAC\cite{sohn2020simple}                          &                                                    & 77.45          & 44.64          \\
Instant Teaching\cite{zhou2021instant}              &                                                     & 78.3           & 48.7           \\
ISMT\cite{yang2021interactive}                          &                                                & 77.2           & 46.2           \\
Multi Phase\cite{wang2021data}                   &                                                     & 77.4           & -              \\
Unbiased Teacher\cite{liu2021unbiased}              &                                                 & 77.4           & 48.7           \\
MUM(Ours)                     &                                                                 & \textbf{78.94} & \textbf{50.22} \\ \hline
CSD\cite{jeong2019consistency}                           & \multirow{6}{*}{\begin{tabular}[c]{@{}c@{}}VOC12\\ +\\ COCO20cls\end{tabular}} & 75.1           & -              \\
STAC\cite{sohn2020simple}                          &                                              & 79.08          & 46.01          \\
Instant Teaching\cite{zhou2021instant}              &                                            & 79.0           & 49.7           \\
ISMT\cite{yang2021interactive}                          &                                & 77.75          & 49.59          \\
Unbiased Teacher\cite{liu2021unbiased}              &                                    & 78.82          & 50.34          \\
MUM(Ours)                           &                                                 & \textbf{80.45}           & \textbf{52.31}          \\ \hline
\end{tabular}
}
\label{tab:voc}
\end{table}


\section{Experiments}
\label{sec:experiments}
\noindent\textbf{Datasets.} We evaluate our proposed method on two standard object detection datasets, PASCAL VOC\cite{everingham2010pascal} and MS-COCO\cite{lin2014microsoft}, following the dominant benchmark of previous SSOD works\cite{jeong2019consistency, sohn2020simple, liu2021unbiased, zhou2021instant}.
The benchmark has three protocols: (1) COCO-Standard: we randomly select 0.5, 1, 2, 5, and 10\% of COCO2017-train dataset as labeled training data and treat the remaining data as unlabeled training data. 
(2) COCO-Additional: we utilize whole COCO2017-train dataset as labeled training data and the additional COCO2017-unlabeled dataset as the unlabeled training data.
(3) VOC: we use VOC07-trainval set as the labeled training data and VOC12-trainval set as the unlabeled training data. To investigate the effect of the increased unlabeled data, we use COCO20cls\cite{jeong2019consistency} as additional unlabeled data.
Model performance is tested on COCO2017-val and VOC07-test for evaluation following STAC\cite{sohn2020simple} and Unbiased-Teacher\cite{liu2021unbiased}.

\noindent\textbf{Implementation Details.}
We use Faster-RCNN\cite{ren2015faster} with FPN\cite{lin2017focal} and ResNet-50\cite{he2016deep} initialized by ImageNet\cite{deng2009imagenet} feature extractor as base network architecture following Unbiased-Teacher\cite{liu2021unbiased}.
We use training schedules of 180K, 360K, 45K and 90K iterations for COCO-Standard, COCO-Additional, VOC, and VOC with COCO20cls. 
Other training configuration is same as Detectron2 \cite{wu2019detectron2} and Unbiased-Teacher\footnote{Code : \url{https://github.com/facebookresearch/unbiased-teacher}} for the sake of fair comparison.
We use a low initial decay rate $\delta=0.5$ and gradually increase it to 0.9996 at the same step of burn-in stage used in the baseline\cite{liu2021unbiased} instead of employing burn-in stage. 
MUM has its own two hyperparameters: $N_G$, $N_T$, which are the number of images to form a group and the number of tiles in each image axis, respectively. We use $N_G = N_T =4$, which were found in our ablation study.

\subsection{Results}
\noindent\textbf{MS-COCO.}
We first evaluate our proposed method on MS-COCO dataset with two protocols, COCO-Standard and COCO-Additional.
As shown in Table \ref{tab:coco}, our approach obtains $\sim$2\%p mAP gain over the baseline\cite{liu2021unbiased} and surpasses all of the recent state-of-the-art results.
Specifically, for the 0.5\% protocol in Table \ref{tab:coco}, MUM achieves 18.54\% mAP which improves 11.71\%p over the supervised results, and its performance is comparable to Instant Teaching and ISMT in 1\% protocol as well (18.05 and 18.88).
MUM brings more improvements when the labeled data is scarce (COCO-Standard 0.5\% and 1\%) since it generates many training samples with natural occlusions and diverse appearances.

\noindent\textbf{Pascal VOC.}
We next test the proposed MUM on the Pascal VOC dataset with two protocols in Table \ref{tab:voc}.
As in MS-COCO, our method consistently outperforms the state-of-the-art methods and achieves 1$\sim$2\%p mAP improvement over the baseline in both AP$_{50}$ and AP$_{50:95}$.
Specifically, MUM has 7.82\%p, 10.18\%p improvement for AP$_{50}$ and AP$_{50:95}$ respectively, over the supervised baseline. 
While Unbiased Teacher shows relatively weak competitiveness compared to the other researches in the VOC dataset unlike the above COCO results, our method still surpasses the other state-of-the-art results with a large margin.
These results demonstrate that our proposed method, MUM, can improve the existing SSOD consistently in various datasets.
\subsection{Ablation Study}
\noindent\textbf{Analysis of $N_G$ and $N_T$.} 
MUM needs two hyperparameters: $N_G$ and $N_T$, which indicate the  number of images to group and shuffle the tiles, and the number of tiles in each image axis.
In order to investigate the effect of two hyperparameters, we examine the performance of MUM with $N_T \in \{2,4,8,16\}$ and $N_G \in \{2,4,6,12\}$ in Table \ref{tab:ng_nt}.
We find $N_G = N_T = 4$ is an appropriate choice to keep MUM with diverse appearances and semantic information without much loss in the geometric information.
When $N_T$ increases to 8 and 16, the performances drop drastically since the tile's size becomes too small to keep the semantic information of positive objects.

We also observe the performance drop with increased $N_G$. However, it was negligible compared to the $N_T$ case.
Especially when $N_G$ further increases beyond 4, AP$_{50:95}$ decreases slightly, but AP$_{50}$ increases a bit.
We assume this phenomenon is because large $N_G$ encourages the network to distinguish objectness and classify objects better by using more occluded images (AP$_{50}$ increased), while it prohibits the network from getting more accurate bounding box position (AP$_{50:95}$ decreased). However, the performance differences are not significant.

\noindent\textbf{Swin Transformer Backbone.} 
To further investigate the generality of MUM, we replace ResNet with Swin Transformer\cite{liu2021swin} and examine the performance in COCO-Standard protocols (Table \ref{tab:swin}).
We use Swin-T, which is comparable to ResNet-50 in terms of computational complexity, from open-source library timm \cite{rw2019timm}.
We first examine Unbiased-Teacher \cite{liu2021unbiased} baseline with Swin backbone.
We set the EMA decay rate to an empirically-found value, $\delta=0.999$, since the default value (0.9996) brings poor results, even worse than the supervised baseline.
And then, we apply MUM to the baseline configuration.
In every protocols, MUM achieves $\sim1$\%p improvement over the baseline.
The efficacy of MUM in Swin is relatively marginal compared to the CNN since MUM possibly hurts the characteristics of the long-range dependency of Transformer.

\noindent\textbf{Supervised Classification.}
MUM can enjoy a regularization effect without any interpolated label, so we extend this idea to the supervised classification task.
We conducted additional experiments for the ImageNet\cite{deng2009imagenet} classification task under a supervised-learning setting. 
We compared MUM with vanilla ResNet, Cutout, Mixup, and CutMix by following the experimental protocol and training framework of CutMix\footnote{Code : \url{https://github.com/clovaai/CutMix-PyTorch}}.
We unmix the mixed features after layer-1 of ResNet and set $N_G$ and $N_T$ as 4 found in SSOD experiments.

As shown in Table \ref{tab:classification}, MUM outperforms the other methods except for CutMix with a top-1 error rate of 22.39\%, which shows that MUM could also be used as a general data augmentation method for classification task.
Compared to Cutout and Mixup, MUM generates much less information loss on the image, leading to a lower error rate.
Furthermore, there is still room for improvement by finetuning the $N_G$, $N_T$, and layer location for unmixing.

\begin{table}[]
\caption{Comparison of mAP with various values of $N_G$ and $N_T$ in COCO-Standard 1\% protocol. For simplicity, we set the training step, batch size as 45K and 12, respectively. We use fixed random seeds to remove the randomness.}
\label{tab:ng_nt}
\centering
\adjustbox{width=0.8\linewidth}
{
\begin{tabular}{|c|c|c|ccc|}
\hline
Methods              & $N_G$                 & $N_T$                 & $AP_{50:95}$ & $AP_{50}$ & $AP_{75}$  \\ \hline
Baseline             & 1                  & 1                  & 18.40                         & 34.99                     & 17.48 \\ \hline
\multirow{7}{*}{MUM} & \textbf{4}         & \textbf{4}         & \textbf{18.99}               & \textbf{36.09}            & \textbf{18.31} \\ \cline{2-6} 
                     & \multirow{3}{*}{4} & 2                  & 18.52                        & 35.25                     & 17.61 \\
                     &                    & 8                  & 18.28                        & 35.19                     & 17.00    \\
                     &                    & 16                 & \color{red}16.46                        & \color{red}31.93                     & \color{red}15.22 \\ \cline{2-6} 
                     & 2                  & \multirow{3}{*}{4} & 18.92                        & 35.94                     & 17.89 \\
                     & 6                  &                    & 18.85                        & \color{blue}36.27                     & 17.66 \\
                     & 12                 &                    & 18.84                        & \color{blue}36.12                     & 17.56 \\ \hline
\end{tabular}
}
\vspace{-1mm}
\end{table}

\begin{table}
\caption{Comparison of mAP with Unbiased Teacher and MUM with Swin Transformer backbone in COCO-Standard. For simplicity, we set the traning step, batch size as 60K and 16, respectively. We use fixed random seeds to remove the randomness. $^+$ denotes our experiments.}
\centering
\adjustbox{width=0.95\linewidth}
{
\begin{tabular}{|l|ccccc|}
\hline
\multicolumn{1}{|c|}{\multirow{2}{*}{Methods}} & \multicolumn{5}{c|}{COCO-Standard}    \\
\multicolumn{1}{|c|}{}                         & 0.5\% & 1\%   & 2\%   & 5\%   & 10\%  \\ \hline
Supervised                                     & 10.16 & 13.43 & 18.7 & 23.67 & 27.41 \\
Unbiased-Teacher$^+$                               & 15.95 & 19.8 & 24 & 27.88 & 30.48 \\
MUM(Ours)                                      & \textbf{16.52} & \textbf{20.5} & \textbf{24.5} & \textbf{28.35} & \textbf{30.58} \\ \hline
\end{tabular}
}
\label{tab:swin}
\vspace{-1mm}
\end{table}

\begin{table}[]
\caption{Ablation study on COCO-Standard 0.5\% with Swin Transformer. T and T$^*$ denote default teacher ($\delta=0.9996$), and refined teacher ($\delta=0.999$) which is empirically found for Swin backbone. Note that the supervised only AP is 10.16 in Table \ref{tab:swin}.}
\centering
\adjustbox{width=0.9\linewidth}
{
\begin{tabular}{lcccc|r|r}
\hline
                         & \multicolumn{1}{l}{Cutout} & \multicolumn{1}{l}{MUM}  & \multicolumn{1}{l}{T} & \multicolumn{1}{l|}{T$^*$} & Teacher & Student \\ \hline
\multicolumn{1}{l|}{(1)} & \checkmark                 &                                                   & \checkmark            &                          & {\color{red}8.27} & 8.44    \\
\multicolumn{1}{l|}{(2)} & \checkmark                 &                                                    &                       & \checkmark               & 15.95   & 14.68   \\
\multicolumn{1}{l|}{(3)} & \checkmark                 & \checkmark                                         & \checkmark            &                          & 14.55   & 14.22   \\
\multicolumn{1}{l|}{(4)} & \checkmark                 & \checkmark                                         &                       & \checkmark               & \textbf{16.52}   & 15.38  
\end{tabular}
}
\label{tab:teacher_data}
\end{table}

\begin{table}[]
\caption{Experiments results of MUM and existing IR methods, Cutout\cite{devries2017cutout}, Mixup\cite{zhang2017mixup}, and CutMix\cite{cutmix} in supervised classification task.}
\vspace{-1mm}
\centering
\adjustbox{width=0.8\linewidth}
{
\begin{tabular}{|l|cc|}
\hline
Methods   & Top-1 Err (\%) & Top-5 Err (\%) \\ \hline
Baseline  & 23.68          & 7.05           \\
Cutout\cite{devries2017cutout}    & 22.93          & 6.66           \\
Mixup\cite{zhang2017mixup}     & 22.58          & 6.40           \\
CutMix\cite{cutmix}    & \textbf{21.40} & 5.92           \\
MUM(Ours) & \color{blue}22.39 & 6.44           \\ \hline
\end{tabular}
}
\label{tab:classification}
\vspace{-1mm}
\end{table}

\section{Discussion}
\label{sec:discussions}
\noindent\textbf{Teacher and Data augmentation.} 
Building a good teacher and applying effective data augmentation is very important for pseudo-labeling-based SSOD systems, as mentioned in Fig. \ref{fig:ssl_system}.
In order to analyze how the two factors affect an SSOD system, we compare the worse and better approaches for building a teacher and augmenting data in Table \ref{tab:teacher_data}.
(1) With only Cutout (worse augmentation) and default EMA decay rate (worse teacher), the teacher's performance is even worse than the student's (8.44$\to$8.27), and semi-supervised learning rather hurts mAP performance compared to the supervised learning (8.27 vs. 10.16).
(2,3) If either one of MUM (better augmentation) and controlled EMA decay rate (better teacher) is used, semi-supervised learning turns to be helpful.
A better teacher and better augmentation result in 5.79 and 4.39 mAP improvements (10.16 vs. 15.95, 14.55), respectively. 
It is remarkable that (3) still improves performance even with a worse teacher because MUM generates mixed input images which are difficult but worth learning and make SSOD helfpul. 
Finally, using both better teacher and augmentation leads to the best performance (16.52) in (4).
We confirm the importance of building a good teacher and data augmentation strategy in the SSOD framework from the experimental results.

\begin{figure}[t!] 
\includegraphics[width=1.\linewidth]{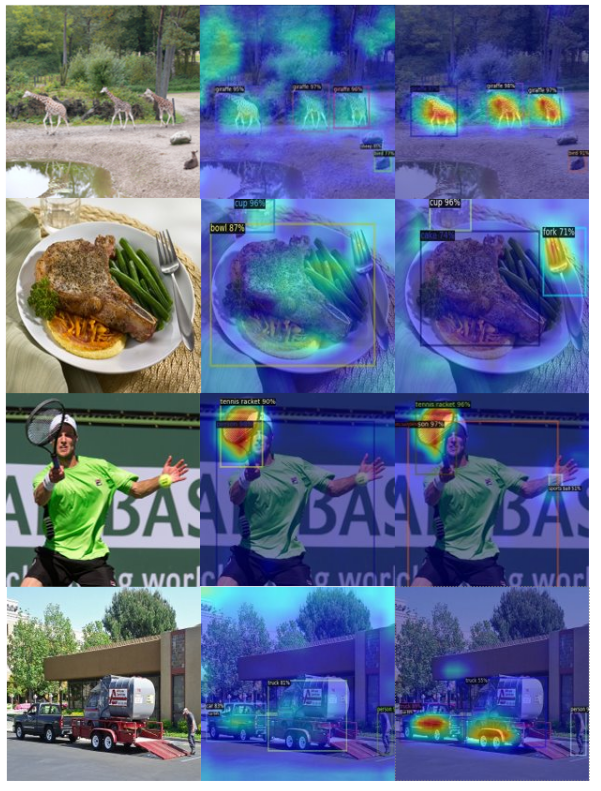}
\caption{Class-Activation-Map (CAM) and box prediction results are provided. 
From left to right, each column shows the original images, outputs of Unbiased-Teacher and MUM, respectively.
From top to bottom, the activated classes of each row are giraffe, fork, sports ball, and truck, respectively.}
\label{fig:cam}
\end{figure}

\noindent \textbf{Class-Activation-Map(CAM) Results.}
We further investigate the superiority of MUM over Unbiased Teacher by comparing the qualitative results of GradCAM\cite{selvaraju2017grad} and box predictions in Fig. \ref{fig:cam}.
We use Faster-RCNN with ResNet-50 and pre-trained weight in COCO-Standard 1\% to get the results.
We find that MUM concentrates more on the local region while baseline tries to look at global features, which allows the network with MUM to find small objects better such as sports ball and fork.
Additionally, MUM classifies trucks and giraffe by highly focusing on each object.
These results show that MUM encourages the network to extract meaningful features in the local region.
Table \ref{tab:size} also provides quantitative results that MUM is more effective on small objects rather than large ones.

\begin{table}
\caption{Comparison of Unbiased Teacher and MUM by various APs in COCO-Standard 1\% protocol.}
\centering
\adjustbox{width=0.95\linewidth}
{
\begin{tabular}{|l|cccc|}
\hline
Methods          & $AP_{50:95}$ & $AP_{S}$  & $AP_{M}$   & $AP_{L}$   \\ \hline
Unbiased Teacher\cite{liu2021unbiased} &  20.70             & 8.93  & 21.85 & 28.07 \\
MUM(Ours)              & 21.81              & \textbf{9.86} & \textbf{23.66} & 27.91\\ \hline
\end{tabular}
}
\label{tab:size}
\end{table}

\begin{figure}[t]
    \centering
    \includegraphics[width=0.85\linewidth]{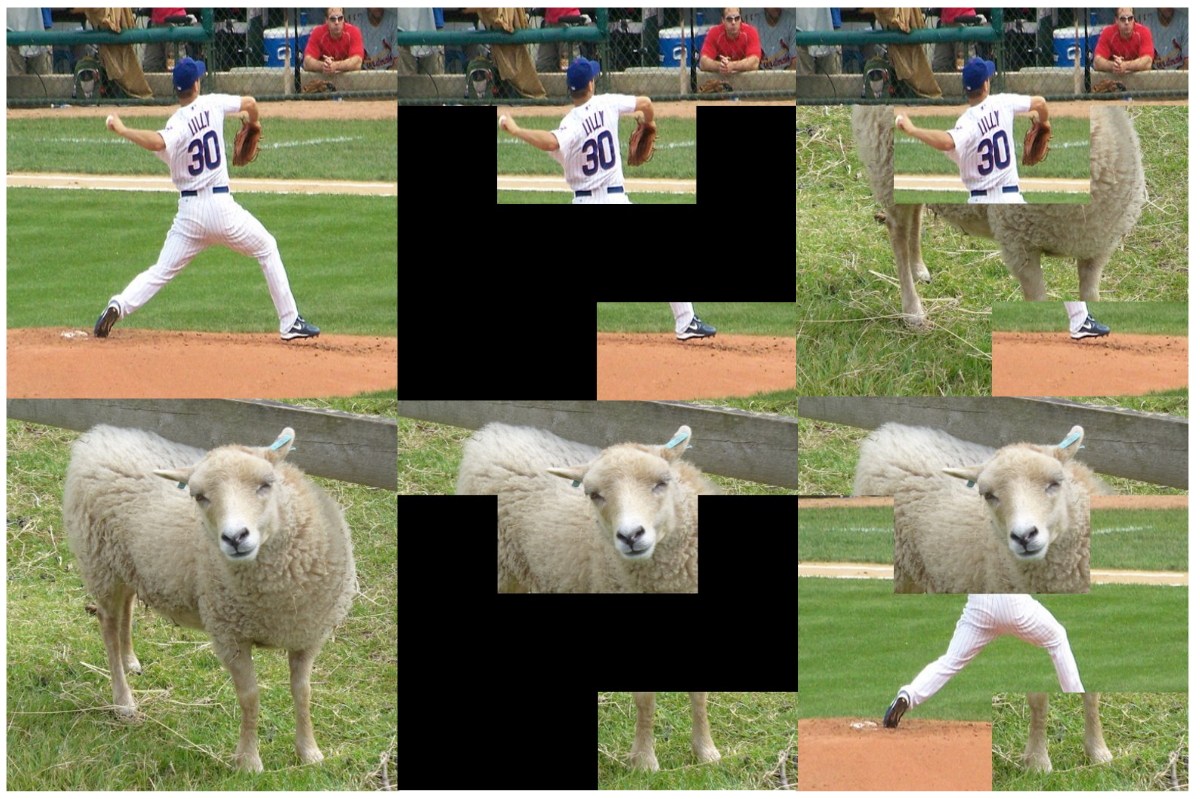}
         \caption{Comparison among original, Cutout, and MUM images. For simplicity and clear comparison, we assume the blocked region of Cutout is the same as mixed region of MUM and set $N_G$ and $N_T$ as 2 and 4, respectively.}
         \label{fig:cutout_tut}
    \vspace{-5mm}
\end{figure}

\noindent\textbf{Connection to Cutout.}
Cutout\cite{devries2017cutout} can be used in semi- and supervised object detection task \cite{bochkovskiy2020yolov4, ge2021yolox,liu2021unbiased} as a strong augmentation method by replacing the pixel blocks with random noisy values and generating diverse appearances and occlusions in training images.
However, the information loss in the image is inevitable since it blocks some areas with noise.
In addition, semantic information of an image that is crucial to predict the correct label can be significantly lost in the worst cases.
On the other hand, our method creates natural occlusion between positive objects similar to Cutout because MUM mixes different images.
However, MUM is able to preserve the semantic information of inputs because it doesn't block the original image with random noise and has a reassembling process in the feature space.
Fig. \ref{fig:cutout_tut} provides examples of augmented images, and shows the difference between Cutout and MUM.
Additionally, we conduct the supervised object detection experiments following the configuration of Detectron2\cite{wu2019detectron2} with Cutout and MUM, and achieve 36.87 and 38.12 mAP, respectively. 
We guess the characteristics of preserving the information of MUM bring the results.

\section{Conclusion}
\label{sec:conclusion}
In this paper, we investigate the pseudo-label-based SSOD system and propose the Mix/UnMix (MUM) data augmentation method, which mixes tiled input images and reassembles feature tiles to generate strongly-augmented images, while preserving the semantic information in the image space.
On top of the pseudo-label-based SSOD framework, MUM obtains consistent performance improvement in SSOD benchmarks and achieves state-of-the-art results.
We extend our experiments to a different backbone, Swin Transformer, and also applied MUM to a supervised ImageNet classification task. The experimental results show that our method is competitive with the existing IR methods  
and can also be used as a general regularization method for general architectures, and general tasks.
We also provide Grad-CAM results to give further evidence why MUM works better. 
Additionally, we analyze the effect of teacher network and data augmentation to properly understand the MUM and SSOD framework.
MUM has a weakness in accurately locating the prediction box since it splits the objects and blinds the edges.
We believe that generating optimized mixing masks using saliency map of objects like \cite{kim2020puzzle, kim2021co} could solve the above problem, and leave it as future work.

\newpage
{\small
\bibliographystyle{ieee_fullname}
\bibliography{egbib}
}

\newpage
\appendix

\section{Training details and stability}
\begin{figure}[h]
    \centering
    \includegraphics[width=1.\linewidth, height = 1.\linewidth]{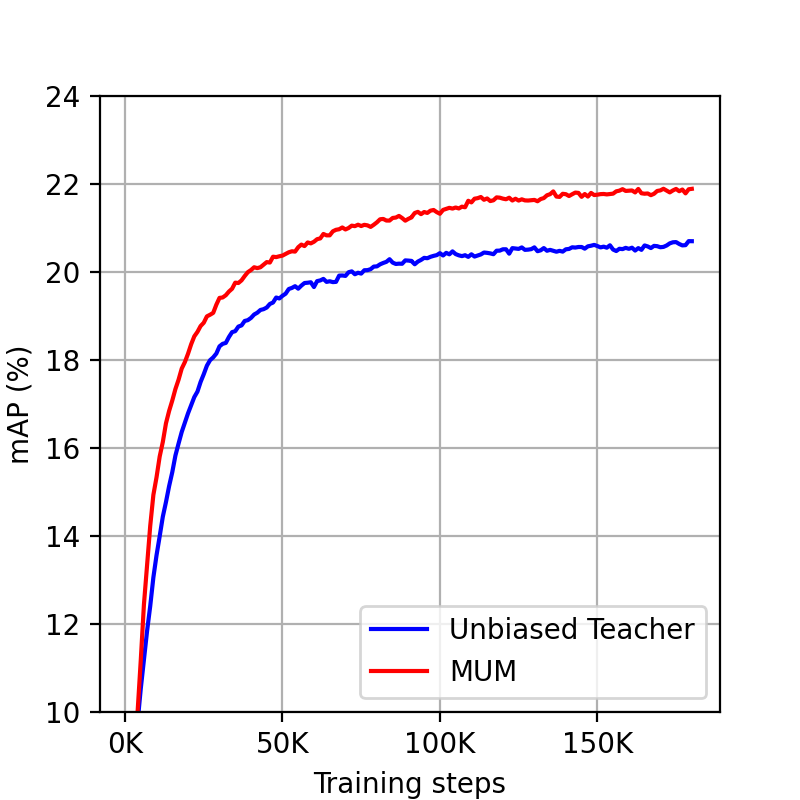}
    \caption{Training curves of the Unbiased Teacher and our MUM. 
    MUM shows a better training curve with no more variance comparing the Unbiased Teacher.}
    \label{fig:training_curve}
\end{figure}

In this section, we provide training details of hyperparameters used in our experiments, as show in Table. \ref{tab:hyperparam}. 
The most of parameters are from the Unbiased Teacher for the sake of fair comparison.
since MUM is easy to add to any framework and prevent losing semantic information, our training process based on the Unbiased Teacher [27] was stable, and the training accuracy curve rose upward with slight variance  (see Fig. \ref{fig:training_curve}).
Thanks to this, we have conducted our experiments with the default hyperparameters in the Unbiased Teacher [27] and could verify that our MUM works well as an add-on to the baseline SSOD work and achieves better performance in fair comparison.

\section{Relation between $N_T$ and foreground-image ratio}
\begin{figure}[ht]
    \centering
    \includegraphics[width=1.\linewidth, height = 1.\linewidth]{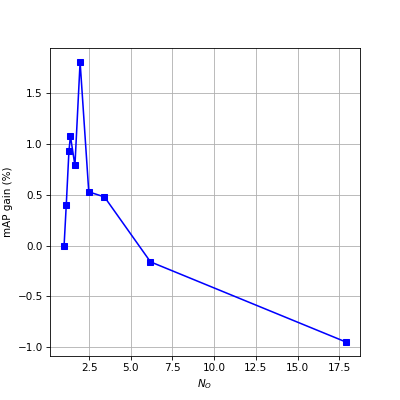}
    \caption{$N_T$ vs. gain of mAP in the COCO validation dataset}
    \label{fig:nt_and_mAP}
\end{figure}

The number of tiles ($N_T$) correlates with the foreground-image ratio, because the degree of splitting and occluding the foreground object is simultaneously related to the foreground object size and tile size.
In order to further investigate the correlation between foreground-image ratio and $N_T$, here we define a new parameter, $N_O$ as the average number of tiles where a single foreground object lies.
The relationship between the gain of mAP and $N_O$ with various $N_T$ and object size in the COCO validation dataset is summarized in Table \ref{tab:nt_andmAP} and Fig. \ref{fig:nt_and_mAP}

We find $1.2 \le N_O \le 2.5$ is an acceptable range for the AP gain. 
It is reasonable that too small $N_O$ ($<1.2$) means that augmentation is not enough, while too large $N_O$ ($>2.5$) tells that the foreground object is teared into too many pieces.
This explains our choice of the tile size, $N_T$=4, was a reasonable.
The above experiment implies that $N_T$ should be adjusted for the image and foreground object's resolution depending on the $N_O$ and MUM can be enhanced by sophisticatedly generating the mixing mask.

\begin{table}[hb]
\caption{The relationship between $N_O$ and AP with varying $N_T$ and object size.
}
\begin{center}
\begin{adjustbox}{width=1.0\linewidth}
\begin{tabular}{|clll|}
\hline
\multicolumn{4}{|c|}{$N_O$ / AP / AP gain}                                 \\ \hline
\multicolumn{1}{|c|}{$N_T$} & \multicolumn{1}{c}{Small} & \multicolumn{1}{c}{Medium} & \multicolumn{1}{c|}{Large} \\ \hline
\multicolumn{1}{|l|}{1 (baseline)}     & 1.00/8.93/0      & 1.00/21.85/0          & 1.00/28.07/0         \\
\multicolumn{1}{|l|}{2}     & 1.13/9.33/+0.40  & 1.38/22.93/+1.08 & 2.48/28.60/+0.53  \\
\multicolumn{1}{|l|}{4}     & 1.29/9.86/+0.93 & 1.96/23.66/\color{blue}{+1.81} & 6.19/27.91/-0.16 \\
\multicolumn{1}{|l|}{8}     & 1.65/9.72/+0.79 & 3.42/22.33/+0.48 & 17.9/27.12/\color{red}{-0.95} \\ \hline
\end{tabular}
\end{adjustbox}
\label{tab:nt_andmAP}
\end{center}
\end{table}

\begin{table*}[ht]
\caption{Hyperparameters for various protocols}
\centering
\begin{adjustbox}{width=1.0\linewidth}
\begin{tabular}{|l|c|c|c|c|c|c|}
\hline
\multicolumn{1}{|c|}{Hyperparameters} & \multicolumn{1}{|c|}{Description} & \multicolumn{1}{|c|}{COCO-Standard} & \multicolumn{1}{|c|}{COCO-Additional} & 
\multicolumn{1}{|c|}{VOC12} & \multicolumn{1}{|c}{VOC12+COCO20cls} & \multicolumn{1}{|c|}{Swin} \\ \hline
$\tau$ & Confidence threshold of pseudo label & 0.7 & 0.7 & 0.7 & 0.7 & 0.7 \\
$\lambda_u$     & Unsupervised loss weight      & 4.0 & 2.0 & 4.0 & 4.0 & 4.0     \\
$\delta$    &   EMA decay rate  & 0.9996 & 0.9996 & 0.9996 & 0.9996 & 0.999 \\
$p$   & Percentage of applying MUM & 1.0 & 1.0 & 1.0 & 1.0 & 1.0 \\
Optimizer & Training optimizer & SGD & SGD & SGD & SGD & SGD \\
LR     & Initial learning rate & 0.01 & 0.01 & 0.01 & 0.01 & 0.01 \\ 
Momentum     & SGD momentum & 0.9 & 0.9 & 0.9 & 0.9 & 0.9 \\ 
Weight Decay     & Weight decay & 1e-4 & 1e-4 & 1e-4 & 1e-4 & 1e-4 \\ 
Training Steps     & Total training steps & 180K  & 360K  & 45K  & 90K & 60K \\ 
Batch Size     & Batch size & 32 & 32 & 32 & 32 & 16 \\ 
\hline
\end{tabular}
\end{adjustbox}
\label{tab:hyperparam}
\end{table*}

\end{document}